\documentclass[letterpaper]{article} % DO NOT CHANGE THIS
\usepackage{aaai2026}  % DO NOT CHANGE THIS
\usepackage{times}  % DO NOT CHANGE THIS
\usepackage{helvet}  % DO NOT CHANGE THIS
\usepackage{courier}  % DO NOT CHANGE THIS
\usepackage[hyphens]{url}  % DO NOT CHANGE THIS
\usepackage{graphicx} % DO NOT CHANGE THIS
\usepackage{siunitx}
\usepackage{booktabs}
\usepackage{amsmath}
\usepackage{xcolor}
\usepackage{natbib}
\usepackage{natbib}  % DO NOT CHANGE THIS AND DO NOT ADD ANY OPTIONS TO IT
\usepackage{caption} % DO NOT CHANGE THIS AND DO NOT ADD ANY OPTIONS TO IT
\nocopyright

\usepackage{diagbox}
\usepackage{bbding}
\usepackage{algorithm}
\usepackage{algpseudocode}
\usepackage{multirow}
\usepackage{makecell}
\usepackage{graphicx}
\usepackage{amsmath}
\usepackage{amssymb}
\usepackage{booktabs}
\usepackage{xcolor}
\usepackage{colortbl}
\usepackage{upgreek}

\usepackage[most]{tcolorbox} % 引入 tcolorbox 宏包

\newtcolorbox{promptbox}{
    colback=cyan!5!white,        % 更柔和的淡蓝背景（比 blue!10 更贴近图中效果）
    colframe=cyan!55!black,      % 浅蓝灰色边框
    fonttitle=\bfseries,     
    coltitle=white,          
    title=Prompt:,           
    arc=1mm,                 
    boxrule=1pt, 
    width=7.5cm, % 设置固定宽度为 10 厘米
}

\usepackage{newfloat}
\usepackage{listings}
\DeclareCaptionStyle{ruled}{labelfont=normalfont,labelsep=colon,strut=off} % DO NOT CHANGE THIS
\floatstyle{ruled}
\newfloat{listing}{tb}{lst}{}
\floatname{listing}{Listing}
\title{CoDe-NeRF: Neural Rendering via Dynamic Coefficient Decomposition}
\author{
    Wenpeng Xing\textsuperscript{\rm 1},
    Jie Chen\textsuperscript{\rm 2},
    Zaifeng Yang\textsuperscript{\rm 3},
    Tiancheng Zhao\textsuperscript{\rm 4},
    Gaolei Li\textsuperscript{\rm 5},
    Changting Lin\textsuperscript{\rm 4},
    Yike Guo\textsuperscript{\rm 6},
    Meng Han\textsuperscript{\rm 1}
}
\affiliations{
    \textsuperscript{\rm 1}Zhejiang University\quad
    \textsuperscript{\rm 2}Hong Kong Baptist University\quad
    \textsuperscript{\rm 3}A*STAR, Singapore\quad
    \textsuperscript{\rm 4}Binjiang Institute of Zhejiang University\quad
    \textsuperscript{\rm 5}Shanghai Jiao Tong University\quad
    \textsuperscript{\rm 6}Hong Kong University of Science and Technology
}

\newcommand{\Tensor}{\mathcal{T}}
\newcommand{\Vectorss}{\mathbf{v}}
\newcommand{\Matrix}{\mathbf{M}}

\newcommand{\ignore}[1]{}

\newcommand{\Dens}{\sigma}

\newcommand{\Step}{\Delta}

\definecolor{colorfirst}{rgb}{.866,.945, 0.831} % green
\definecolor{colorsecond}{rgb}{1, 0.98, 0.83} % yellow
\definecolor{colorthird}{rgb}{1, 1, 1} % white
\usepackage{bibentry}

\begin{document}

\maketitle

\begin{abstract}

Neural Radiance Fields (NeRF) have shown impressive performance in novel view synthesis, but challenges remain in rendering scenes with complex specular reflections and highlights. Existing approaches may produce blurry reflections due to entanglement between lighting and material properties, or encounter optimization instability when relying on physically-based inverse rendering. In this work, we present a neural rendering framework based on {dynamic coefficient decomposition}, aiming to improve the modeling of view-dependent appearance. Our approach decomposes complex appearance into a shared, static {neural basis} that encodes intrinsic material properties, and a set of dynamic {coefficients} generated by a {Coefficient Network} conditioned on view and illumination. A {Dynamic Radiance Integrator} then combines these components to synthesize the final radiance. Experimental results on several challenging benchmarks suggest that our method can produce sharper and more realistic specular highlights compared to existing techniques. We hope that this decomposition paradigm can provide a flexible and effective direction for modeling complex appearance in neural scene representations.

\end{abstract}

\section{Introduction}
Creating photorealistic, freely-navigable 3D scenes from a sparse set of 2D images is a fundamental challenge in computer graphics and vision, with critical applications in virtual reality (VR), augmented reality (AR), and digital twins. The advent of Neural Radiance Fields (NeRF) and its successors has revolutionized high-quality novel view synthesis by representing scenes as differentiable volumetric functions. However, while these methods excel at rendering diffuse or Lambertian surfaces, they still face significant challenges with the glossy or non-Lambertian surfaces ubiquitous in the real world, which exhibit complex highlights and specular reflections. These high-frequency, view-dependent effects demand a highly expressive model to be captured accurately, a weak point for many current leading methods.

Existing approaches have largely followed two paths to address this problem, each with its own limitations. The first category of methods, including TensoRF and 3D Gaussian Splatting (3DGS), focuses on improving the training and rendering efficiency of NeRF. Despite their remarkable success in speed, they often "bake" material and illumination effects into a unified appearance representation. This entangled representation hinders their ability to render sharp, dynamically changing specular highlights, often resulting in blurry or view-inconsistent artifacts. The second category attempts to disentangle the physical properties of the scene via neural inverse rendering, for instance, by decomposing appearance into diffuse and specular components based on physical BRDF models. However, these methods often rely on complex optimization processes that are prone to unstable local minima, leading to ambiguous decompositions of material and light, and typically require strong geometric or illumination priors.

To overcome these limitations, we propose in this paper a novel neural rendering framework centered on a {dynamic coefficient decomposition}. Our core idea is to break down the complex task of modeling view-dependent appearance into two parts: a static, shared {neural basis} that represents the scene's intrinsic material properties, and a set of dynamic {coefficients} that adaptively change based on the viewing direction and illumination conditions. As illustrated in Figure~\ref{fig:pip}, we design a {Coefficient Network} to generate these dynamic coefficients, which are then non-linearly combined with the basis by a {Dynamic Radiance Integrator} to synthesize the final color. This design avoids the rigid constraints and optimization challenges of traditional physical models, while being more expressive than simple linear models, thus enabling the efficient and high-fidelity rendering of sharp, detailed specular effects.

Our main contributions can be summarized as follows:
\begin{itemize}
    \item We propose a novel dynamic coefficient decomposition framework, which provides a powerful and flexible paradigm for modeling complex view-dependent effects by combining a static neural basis with dynamic modulating coefficients.
    \item We design an effective Coefficient Network that leverages a FiLM-like affine transformation mechanism to generate coefficients, enabling the precise capture of appearance changes caused by view and illumination variations.
    \item Through extensive experiments on multiple challenging benchmarks, we demonstrate that our method achieves state-of-the-art performance in rendering glossy and reflective scenes, significantly outperforming prior methods in both qualitative and quantitative comparisons.
\end{itemize}

\section{Related work} \label{relatedwork}

\subsection{Neural Representations for Novel View Synthesis}
Novel View Synthesis (NVS) aims to generate images of a scene from unobserved viewpoints, given a set of known images. Traditional methods heavily rely on explicit 3D geometry obtained from Multi-View Stereo (MVS)~\cite{schoenberger2016mvs} or Photometric Stereo (PS)~\cite{chen2018ps}, and synthesize new views by warping pixels between views. However, these methods often falter when dealing with specular reflections or textureless regions due to inaccurate depth estimation. Light field methods~\cite{xing2022scale} enable high-quality interpolation within small baselines but struggle to support free-viewpoint rendering over large baselines.

Recent years have seen a breakthrough with implicit neural representations, most notably Neural Radiance Fields (NeRF)~\cite{mildenhall2020nerf}. NeRF maps a 3D coordinate and a 2D viewing direction to a volumetric density and color using a multilayer perceptron (MLP), and leverages a differentiable volumetric rendering equation for end-to-end optimization. This approach has achieved unprecedented photorealistic rendering quality. However, the original NeRF suffers from several limitations, including slow training and rendering speeds, poor generalization, and insufficient capacity to model view-dependent effects like specular highlights.

To address the efficiency bottleneck, the research community has proposed various acceleration strategies. Some works employ explicit data structures like sparse voxel grids~\cite{fridovich2022plenoxels, sun2022direct}, or replace the implicit MLP with highly-efficient queryable feature fields via tensor decompositions~\cite{Chen2022ECCV, fridovich2023k}, significantly reducing training time. Among these, TensoRF~\cite{Chen2022ECCV} models the scene using tensor factorization, striking a favorable balance between efficiency and quality. More recently, 3D Gaussian Splatting (3DGS)~\cite{kerbl3Dgaussians} proposes using 3D Gaussians as an explicit scene representation, enabling real-time, high-quality rendering. However, it can introduce artifacts when rendering complex reflections and dense scenes~\cite{kwon2025r3evision}. Despite these advances in speed and quality, most of these methods still entangle illumination and material properties, limiting their performance in scenes with complex lighting and shiny materials.

\subsection{Relightable and Reflective Scene Modeling}
To more realistically render shiny and reflective scenes, the core challenge lies in disentangling the scene's intrinsic properties (e.g., material, geometry) from extrinsic illumination—a process often referred to as neural inverse rendering. Early works like NeRF-W~\cite{martin2021nerf} introduced transient latent codes for each image to handle dynamic objects and lighting variations, but this acts as a non-physical correction and lacks a true decoupling of scene properties.

To achieve a more physically-accurate decomposition, subsequent works have shifted towards explicitly modeling materials and illumination. These methods typically decompose the scene's radiance into diffuse and specular components. For instance, Ref-NeRF~\cite{verbin2022ref} introduced feature queries along the reflection direction and regularization on surface normals, significantly improving its ability to model specular highlights. PhySG~\cite{zhang2021physg} and NeRO~\cite{liu2023nero} go a step further by integrating physically-based rendering (PBR) models, such as the Bidirectional Reflectance Distribution Function (BRDF), into the neural field, while simultaneously predicting spatially-varying material parameters (e.g., albedo, roughness, metallic) and the environmental illumination. TensoIR~\cite{Jin2023TensoIR} combines this idea with tensor factorization for efficient inverse rendering. However, these methods often rely on complex multi-stage optimization pipelines or strong priors (e.g., precise camera poses, light probes) and are prone to getting trapped in local minima during optimization, leading to an ambiguous decomposition of material and illumination.

In contrast to the works above, our method effectively captures complex view- and illumination-dependent effects through a dynamic, coefficient-driven decomposition. We neither rely on complex physical BRDF models and their unstable inverse optimization, nor are we constrained by the linear expressiveness of traditional basis learning. Through a non-linear fusion of adaptive coefficients and invariant neural bases, our method achieves high-fidelity rendering of specular reflections and highlights at a low computational overhead, without requiring dense views or additional priors, surpassing baseline models such as NeRF, TensoRF, and 3DGS in visual quality.

\begin{figure*}[t]
    \centering
    \includegraphics[width=\linewidth]{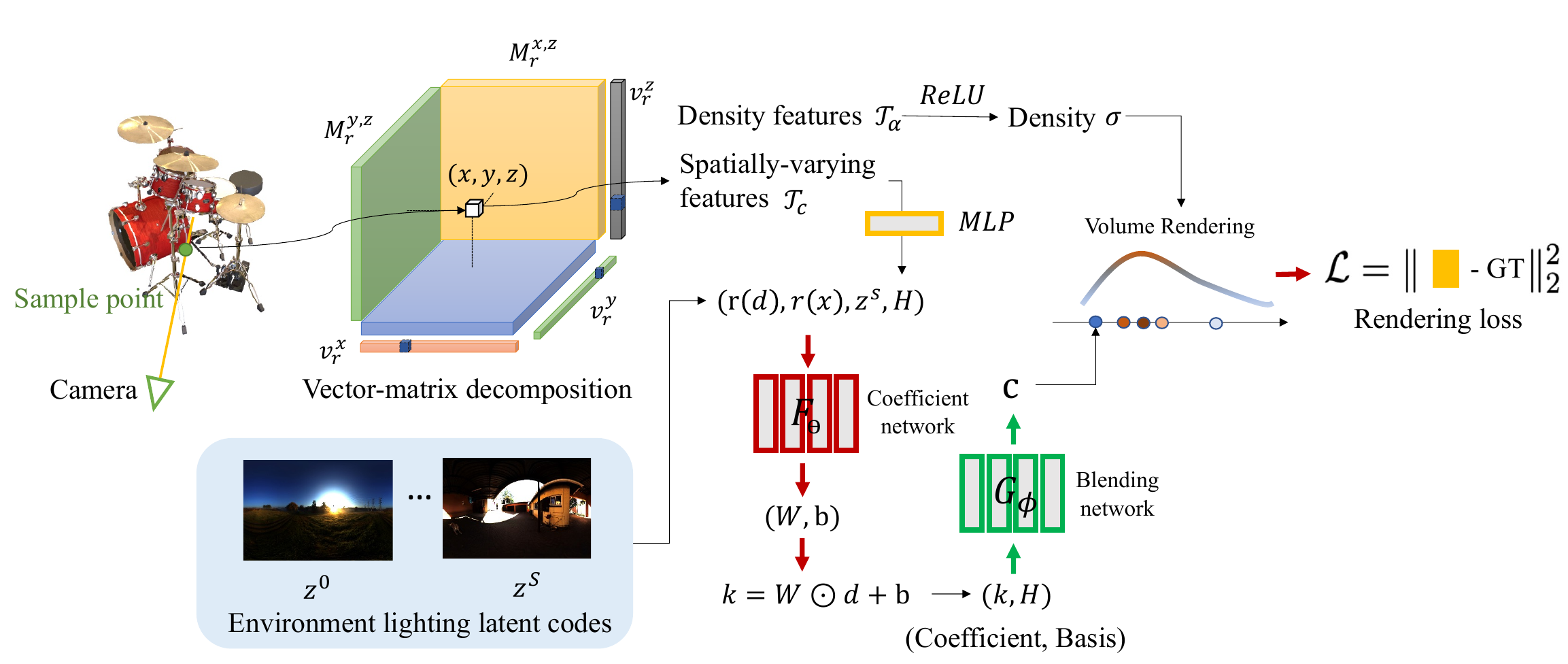}
    \caption{
    Overall pipeline of our method, centered on dynamic coefficient decomposition.
    This approach decomposes scene appearance into a static {neural basis $H$} (encoding material) and a set of dynamic {coefficients $k$} (encoding view and light). A {Coefficient Network $F_\theta$} generates $k$, which are then non-linearly combined with $H$ by a {Dynamic Radiance Integrator $G_\phi$} to produce the final color. This disentangled design enables efficient, high-fidelity rendering of complex reflections.
    }
    \label{fig:pip}
\end{figure*}

\section{Preliminaries}
\subsection{Tensorial Radiance Fields}

TensoRF~\cite{Chen2022ECCV} proposes a compact volumetric representation using vector-matrix (VM) decomposition. For a spatial point \( x = (X, Y, Z) \) in volume \( \mathcal{V} \), density features \( \Tensor_\sigma(x) \) and appearance features \( \Tensor_c(x) \) are queried, supporting efficient gradient propagation during training.

The radiance field tensor \( \Tensor_{\sigma, c} \) is factorized as a sum of outer products:

\begin{equation}
    \Tensor_{\sigma, c} = \sum_{r=1}^{R} \sum_{m \in \{X,Y,Z\}} \Vectorss_r^m \otimes \Matrix_r^{\{X,Y,Z\} \setminus m},
    \label{eqn:vmr}
\end{equation}

where \( \otimes \) is the outer product, and \( R \) is the decomposition rank, enabling low-memory storage and computation. Features at \( x \) are reconstructed via:

\begin{equation}
    \Tensor_{\sigma, c}(x) = \sum_{r=1}^{R_{\sigma, c}} \sum_{m \in \{X,Y,Z\}} \mathcal{A}^m_r(x),
    \label{eqn:vmr_2}
\end{equation}

where \( \mathcal{A}^m_r(x) \) is the mode-specific component along axis \( m \). Related work, such as $K$-Planes~\cite{fridovich2023k}, extends this to planar factorization for dynamic scenes.

Novel views are synthesized using differentiable volume rendering~\cite{mildenhall2020nerf}, integrating radiance along camera rays:

\begin{align}
        \mathbf{C} &= \sum_{x=1}^{X} T_x \left(1 - \exp(-\Dens_x \Step_x)\right) L_o(x, d), \\
        \quad T_x &= \exp\left(-\sum_{j=1}^{x-1} \Dens_j \Step_j\right),
    \label{eq:raymarching}
\end{align}
where \( \Dens_x = \text{ReLU}(\Tensor_\sigma(x)) \) is the density, \( \Step_x \) is the step size, and \( L_o(x, d) \) is the view-dependent color from \( \Tensor_c(x) \).

\section{Proposed Method}

\subsection{Overview}
We propose a novel framework for novel view synthesis that redefines the modeling of view- and illumination-dependent effects through a dynamic coefficient decomposition. Our core innovation lies in a coefficient network \( F_\theta \) that adaptively generates view-conditioned coefficients \( k \), enabling precise capture of complex specular reflections and lighting variations from multi-view images under consistent lighting \( \{\mathcal{I}_i\}_{i=1}^{N} \) or diverse conditions \( \{\mathcal{I}_i^s\}_{i=1,s=1}^{N,S} \). This approach, depicted in Figure~\ref{fig:pip}, achieves unparalleled rendering fidelity by blending these coefficients with a shared neural basis, advancing beyond traditional Neural Radiance Fields \citep{mildenhall2020nerf}.

Specifically, our method represents the scene via a volumetric field \( \mathcal{V} \), where each point \( x \in \mathbb{R}^3 \) is associated with:
\begin{itemize}
    \item Density features \( \Tensor_\sigma(x) \in \mathbb{R}^{R^1_{\sigma} + R^2_{\sigma} + R^3_{\sigma}} \), governing geometry,
    \item Appearance features \( \Tensor_c(x) \in \mathbb{R}^{R^1_{c} + R^2_{c} + R^3_{c}} \), encoding material properties.
\end{itemize}
These appearance features are projected into a compact set of view-agnostic neural bases \( H = \{ h_n \}_{n=1}^{N_p} \), which encapsulate reusable reflection patterns. The coefficient network \( F_\theta \) processes \( \Tensor_c(x) \), viewing direction \( d \), and a learnable illumination embedding \( z^s \in \mathbb{R}^D \) to produce coefficients \( k \), which are blended with \( H \) via a Dynamic Radiance Integrator \( G_\phi \) to yield outgoing radiance \( L_o(x, d, z^s) \) for volume rendering. Auxiliary illumination embeddings \( Z = \{ z^s \}_{s=1}^{S} \) enhance flexibility across lighting conditions.

In summary, our framework synergizes the coefficient network \( F_\theta \) for dynamic view-light modulation, the neural basis \( H \) for reusable reflection patterns, the Dynamic Radiance Integrator \( G_\phi \) for non-linear radiance synthesis, and illumination embeddings \( Z \) for versatile lighting adaptation. This integrated design enables high-fidelity rendering for applications like immersive VR.

\subsection{Neural Approximation of Physics-Driven Rendering}

Traditional rendering employs image-based lighting (IBL) to model surface appearance, formalized as:
\begin{align}\label{eq:brdf}
    L_o({x},d) = \int_\Omega f_r({x}, {\omega}_i,d)L_i({x},{\omega}_i)({\omega}_i \cdot \mathbf{n}) \,d{\omega}_i,
\end{align}
where \( L_o(x,d) \) is the outgoing radiance, \( f_r \) the BRDF, \( \mathbf{n} \) the surface normal, and \( \omega_i \) the incoming light direction. This integral, while accurate, demands intensive computation due to dense sampling across the light field.

Such computational complexity limits real-time applications. Previous methods like Neural-PIL \citep{boss2021neural} mitigate this by precomputing shadowing and Fresnel effects into textures, yet incur prolonged training times. Similarly, Spherical Gaussians \citep{boss2021nerd,gardner2019deep,li2020inverse} often fail to capture high-frequency specular details \citep{munkberg2022extracting}. To address these shortcomings, we propose a neural approximation that replaces explicit integration with a dynamic coefficient-driven decomposition.
Specifically, the decomposition consists of the following stages:

\begin{enumerate}
\item {Dynamic Coefficient Prediction.}
The coefficient network \( F_\theta \) generates view- and lighting-adaptive coefficients \( k \):
\begin{equation}\label{eq:coefficient_brdf}
    F_\theta: (r(x), r(d), z^s, H) \rightarrow k,
\end{equation}
where \( r(\cdot) \) denotes positional encoding \citep{mildenhall2020nerf}, and \( H = \{ h_n \}_{n=1}^{N_p} \) is derived from appearance features:
\begin{equation}
    H = \mathcal{W} \cdot \Tensor_c, \quad \mathcal{W} \in \mathbb{R}^{(R^1_c + R^2_c + R^3_c) \times N_p}.
\end{equation}
This stage leverages view direction \( d \) to modulate reflection patterns with high precision.

\item { Neural Blending of Reflectance Patterns.}
The Dynamic Radiance Integrator \( G_\phi \) fuses \( k \) and \( H \) into radiance \( L_o \):
\begin{equation}\label{eq:neural_blending}
    G_\phi: (k, H) \rightarrow L_o.
\end{equation}
This non-linear blending captures complex effects, surpassing static sums. The final radiance is:
\begin{equation}\label{eq:brdf_reformat}
    L_o({x},d) = G_\phi(F_\theta(r(d), r(x), z^s, H), H).
\end{equation}

\item {Illumination Embedding for Flexibility.}
To support varying lighting, we introduce learnable codes \( z^s \in \mathbb{R}^D \), optimized jointly:
\begin{equation}\label{eq:latent_math}
    z^s \approx \int_\Omega L_i(x, \omega_i)(\omega_i \cdot \mathbf{n}) \, d\omega_i.
\end{equation}
Initialized normally, \( z^s \) ensures smooth interpolation across conditions, omitted in single-illumination cases for efficiency.

\end{enumerate}

This illumination embedding complements our core innovation: a dynamic coefficient decomposition that drives view- and lighting-dependent rendering. By integrating \( z^s \) with appearance features and view directions, our framework reformulates the image-based lighting model (Eq.~\ref{eq:brdf}) into a neural approximation, expressed as:
\begin{equation}
    L_o(x,d) = G_\phi(F_\theta(r(d), r(x), z^s, H), H),
\end{equation}
where \( F_\theta \) generates adaptive coefficients \( k \), and \( G_\phi \) blends them with neural bases \( H \) to produce outgoing radiance \( L_o \). This decomposition enables high-fidelity rendering of specular effects, seamlessly bridging physical accuracy with computational efficiency. We next detail the coefficient network \( F_\theta \), which underpins this process by dynamically modulating view-dependent reflection patterns.

\subsubsection{Coefficient Network}

The coefficient network \( F_\theta \), implemented as a multi-layer perceptron (MLP) with dynamic factorization layers, combines point coordinates \( x \), viewing direction \( d \), illumination embedding \( z^s \), and neural basis \( H \) to generate adaptive coefficients \( k \). Inspired by factorized radiance field designs, this network addresses the conditioning challenge \citep{dumoulin2018feature} by decomposing inputs into structured components: \( x \) and \( H \) represent spatial-material features, \( d \) modulates view-dependent effects, and \( z^s \) accounts for lighting variations via tensor contraction. The output \( k \) serves to adaptively weight reflection patterns in \( H \), enabling efficient rendering through low-rank approximation.

To better capture view-dependent effects, we adopt an extended feature-wise linear modulation (FiLM) mechanism with a multi-basis formulation. For a given scene point \( x \), with fixed material basis \( H \) and illumination \( z^s \), the directional input \( d \) is modulated as:
\begin{equation}\label{eq:fea_w_Lin}
    k = W_x \odot d + b_x,
\end{equation}
where \( \odot \) denotes the Hadamard product, \( W_x \in \mathbb{R}^{N_w \times 3} \) is a learned weight matrix, and \( b_x \in \mathbb{R}^{N_w} \) is a bias term. This affine transformation is further constrained by BRDF-inspired priors to improve physical plausibility, particularly in modeling specular highlights and high-frequency details.

Compared to conventional conditioning strategies \citep{chan2021pi, perez2018film, dumoulin2018feature, attal2022learning}, our design offers the following advantages:
\begin{itemize}
    \item \textbf{Structured Input Decomposition:} The use of low-rank factorization allows stable and interpretable modulation of spatial, directional, and lighting inputs.
    \item \textbf{Improved Directional Conditioning:} The FiLM-based formulation helps preserve directional information \( d \) under varying lighting \( z^s \), reducing signal entanglement.
\end{itemize}

This modular and decomposition-aware design of \( F_\theta \) contributes to the overall effectiveness of our dynamic radiance synthesis framework.

\subsubsection{Dynamic Radiance Integrator}

\begin{figure*}[t]
  \centering
  \setlength\tabcolsep{1pt}
  \renewcommand{\arraystretch}{1}

  \begin{minipage}[b]{0.13\linewidth}
    \includegraphics[width=\linewidth]{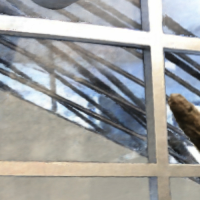}
  \end{minipage}
  \begin{minipage}[b]{0.13\linewidth}
    \includegraphics[width=\linewidth]{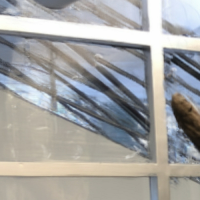}
  \end{minipage}
  \begin{minipage}[b]{0.13\linewidth}
    \includegraphics[width=\linewidth]{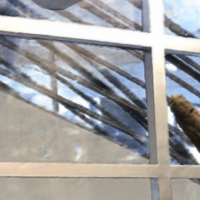}
  \end{minipage}
  \begin{minipage}[b]{0.13\linewidth}
    \includegraphics[width=\linewidth]{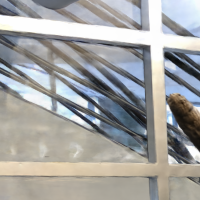}
  \end{minipage}
  \begin{minipage}[b]{0.13\linewidth}
    \includegraphics[width=\linewidth]{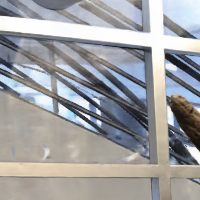}
  \end{minipage}
  \begin{minipage}[b]{0.13\linewidth}
    \includegraphics[width=\linewidth]{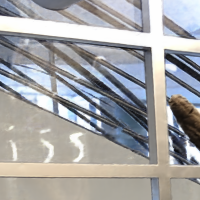}
  \end{minipage}
  \begin{minipage}[b]{0.13\linewidth}
    \includegraphics[width=\linewidth]{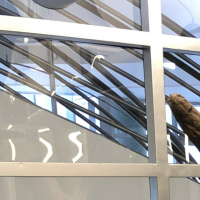}
  \end{minipage}

  \vspace{1pt}

  \begin{minipage}[b]{0.13\linewidth}
    \includegraphics[width=\linewidth]{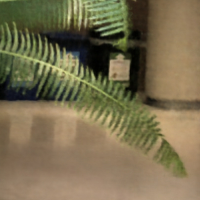}
  \end{minipage}
  \begin{minipage}[b]{0.13\linewidth}
    \includegraphics[width=\linewidth]{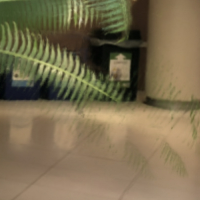}
  \end{minipage}
  \begin{minipage}[b]{0.13\linewidth}
    \includegraphics[width=\linewidth]{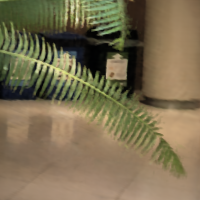}
  \end{minipage}
  \begin{minipage}[b]{0.13\linewidth}
    \includegraphics[width=\linewidth]{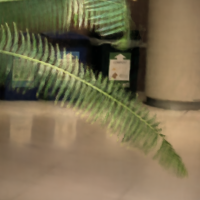}
  \end{minipage}
  \begin{minipage}[b]{0.13\linewidth}
    \includegraphics[width=\linewidth]{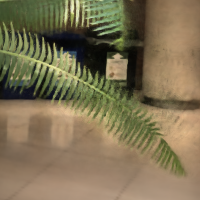}
  \end{minipage}
  \begin{minipage}[b]{0.13\linewidth}
    \includegraphics[width=\linewidth]{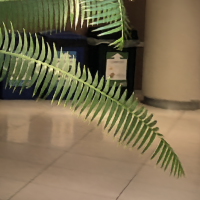}
  \end{minipage}
  \begin{minipage}[b]{0.13\linewidth}
    \includegraphics[width=\linewidth]{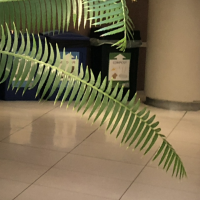}
  \end{minipage}

  % ----------- Fortress Scene -----------
  \vspace{3pt}

  \begin{minipage}[b]{0.13\linewidth}
    \includegraphics[width=\linewidth]{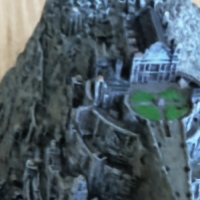}
        \caption*{NeRF}
  \end{minipage}
  \begin{minipage}[b]{0.13\linewidth}
    \includegraphics[width=\linewidth]{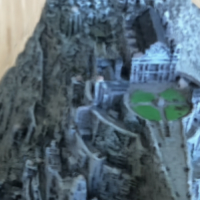}
        \caption*{LLFF}
  \end{minipage}
  \begin{minipage}[b]{0.13\linewidth}
    \includegraphics[width=\linewidth]{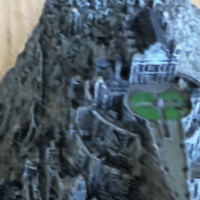}
        \caption*{TensoRF}
  \end{minipage}
  \begin{minipage}[b]{0.13\linewidth}
    \includegraphics[width=\linewidth]{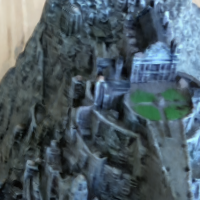}
        \caption*{Nex}
  \end{minipage}
  \begin{minipage}[b]{0.13\linewidth}
    \includegraphics[width=\linewidth]{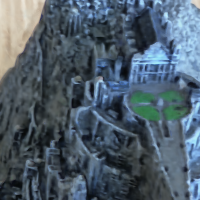}
        \caption*{K-Planes}
  \end{minipage}
  \begin{minipage}[b]{0.13\linewidth}
    \includegraphics[width=\linewidth]{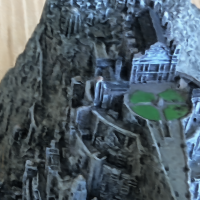}
        \caption*{Ours}
  \end{minipage}
  \begin{minipage}[b]{0.13\linewidth}
    \includegraphics[width=\linewidth]{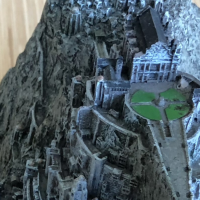}
        \caption*{GT}
  \end{minipage}

  \caption{
    Qualitative comparisons on test views from the Forward-Facing dataset. For each scene (Horns, Fern, Fortress), the full image rendered by our method is shown on the left, followed by close-up crops from various baseline methods and the ground truth.
  }
  \label{fig:llffvisually}
\end{figure*}

The dynamic radiance integrator network \( G_\phi \), implemented as a multi-layer perceptron (MLP), approximates the integral form of image-based rendering (IBR) equations, such as the bidirectional reflectance distribution function (BRDF) integral in Eq. (\ref{eq:brdf}). This network dynamically fuses adaptive coefficients \( k \) with the neural basis \( H \) to synthesize outgoing radiance \( L_o(x, d, z^s) \), enabling high-fidelity capture of view- and illumination-dependent effects, such as specular reflections.

To effectively approximate the complex, non-linear behavior of real-world reflectance, we adopt MLPs due to their superior expressiveness and flexibility. Unlike static weighted sums, MLPs can model high-order interactions between \( k \) and \( H \), outperforming static weighted sums by better approximating non-linear reflectance patterns.
% Mathematical formulation
Mathematically, the network simulates the IBR integral through a decomposed representation. The traditional BRDF integral is:

\begin{equation}
L_o(x, d) = \int_{\Omega} f_r(x, \omega_i, d) L_i(x, \omega_i) (\omega_i \cdot \mathbf{n}) \, d\omega_i.
\end{equation}

By decomposing \( f_r \) into coefficients \( k \) (generated by \( F_\theta \)) and basis \( H \), the approximation becomes:

\begin{equation}
L_o(x, d) \approx G_\phi(k, H) = \sigma \left( W_2 \cdot \sigma(W_1 \cdot [k; H] + b_1) + b_2 \right),
\end{equation}

where \( \sigma \) is an activation function (e.g., ReLU), and \( W_1, W_2, b_1, b_2 \) are learnable parameters. This non-linear fusion captures high-order interactions. Specifically, the MLP concatenates the view-invariant neural basis \( H \) and coefficients \( k \) as input, producing view- and illumination-dependent radiance \( L_o \) as output, as in Eq. (\ref{eq:neural_blending}).

% Collaborative integration
In collaboration with upstream components, \( G_\phi \) integrates seamlessly: coefficients \( k \) from \( F_\theta \) modulate dynamic factors, while \( H \) provides reusable reflectance patterns, and illumination embeddings \( z^s \) ensure adaptability across lighting conditions.  Finally, volume rendering via Eq. (\ref{eq:raymarching}) aggregates \( L_o \) along rays to yield pixel colors \( C \), achieving photorealistic novel views with computational efficiency.

\subsection{Loss Function}

The tensor features \(\Tensor_{c,\sigma}\) and the illumination latent codes \(\{z^s \in \mathbb{R}^{D}\}_{s=1}^{S}\) are jointly optimized by minimizing a combination of an L2 photometric rendering loss and regularization terms, as inspired by tensorial radiance field methods \citep{Chen2022ECCV}. This objective is formulated as:

\begin{equation}
    \mathcal{L} = \|\mathbf{C} - \bar{\mathbf{C}}\|_2^2 + \lambda \cdot \mathcal{L}_{\text{reg}},
\end{equation}

where \(\bar{\mathbf{C}}\) denotes the ground-truth pixel color, and \(\lambda\) is a hyperparameter controlling the strength of the regularization term \(\mathcal{L}_{\text{reg}}\).

To promote spatial smoothness in the appearance features \(\Tensor_c\) and density features \(\Tensor_\sigma\), we employ a Total Variation (TV) regularization loss on the vector-matrix decomposition parameters \(\mathbf{V}\), \(\mathbf{M}\). This regularization mitigates artifacts in sparsely sampled regions and enhances reconstruction stability. The TV loss is expressed as:

\begin{equation}
    \mathcal{L}_{\text{reg}} = \frac{1}{P} \sum \left( \lambda_1 \sqrt{\Delta^2 \mathbf{V}} + \lambda_2 \sqrt{\Delta^2 \mathbf{M}} \right),
    \label{eq:loss_reg}
\end{equation}

where \(P\) is the total number of parameters, \(\Delta^2\) represents the squared differences between neighboring parameters, and \(\lambda_1\), \(\lambda_2\) are weighting hyperparameters.

\section{Experiment Setup} \label{results}

\subsubsection{Implementation Details}

Our method is implemented in PyTorch 1.12. We set an alpha masking threshold of 0.01 to localize surface points, updating the mask at iterations [1000, 2000, 3000, 4000, 5000, 6000, 7000] to exclude empty regions during training.
We employ 3-level tensor decompositions for density and appearance features, with channel dimensions [16, 16, 16] for \(\{R_{\sigma}^1, R_{\sigma}^2, R_{\sigma}^3\}\) and [48, 48, 48] for \(\{R_{c}^1, R_{c}^2, R_{c}^3\}\). Positional encodings for 3D points and view directions use 2 frequency levels. The illumination embedding dimension \(D\) and FiLM modulation dimension \(N_w\) are both set to 32, with 16 neural basis functions (\(N_p = 16\)).
The coefficient network \(F_\theta\) is an 8-layer MLP with 256 hidden units and leaky ReLU activations \citep{xu2015empirical}. The dynamic radiance integrator \(G_\phi\) is a 3-layer MLP with 128 hidden units and ReLU activations.
Training runs for 100k iterations with a batch size of 5,000 and an initial learning rate of 0.02. On a single NVIDIA A100 GPU, training a NeRF-Synthetic scene for 10k iterations takes approximately 2 hours.

\begin{figure}[t]
  \centering
  \begin{minipage}[t]{0.23\linewidth}
    \includegraphics[width=\linewidth]{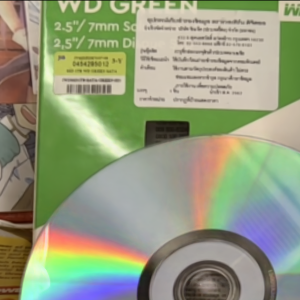}
    \caption*{Nex}
  \end{minipage}
  \hfill
  \begin{minipage}[t]{0.23\linewidth}
    \includegraphics[width=\linewidth]{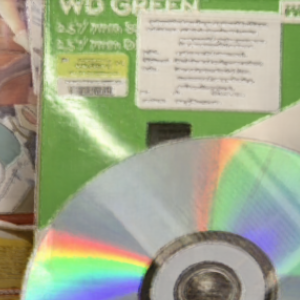}
    \caption*{TensoRF}
  \end{minipage}
  \hfill
  \begin{minipage}[t]{0.23\linewidth}
    \includegraphics[width=\linewidth]{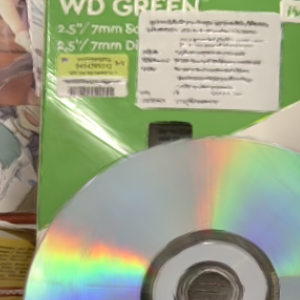}
    \caption*{3D GS}
  \end{minipage}
  \hfill
  \begin{minipage}[t]{0.23\linewidth}
    \includegraphics[width=\linewidth]{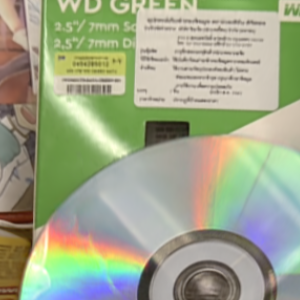}
    \caption*{Plenoxels}
  \end{minipage}
  \hfill
  \begin{minipage}[t]{0.23\linewidth}
    \includegraphics[width=\linewidth]{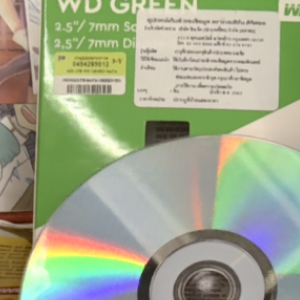}
    \caption*{Ref-NeRF}
  \end{minipage}
  \hfill
  \begin{minipage}[t]{0.23\linewidth}
    \includegraphics[width=\linewidth]{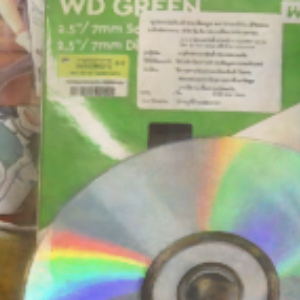}
    \caption*{$K$-Planes}
  \end{minipage}
  \hfill
  \begin{minipage}[t]{0.23\linewidth}
    \includegraphics[width=\linewidth]{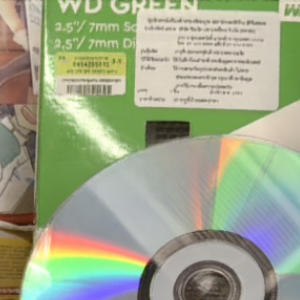}
    \caption*{Ours}
  \end{minipage}
  \hfill
  \begin{minipage}[t]{0.23\linewidth}
    \includegraphics[width=\linewidth]{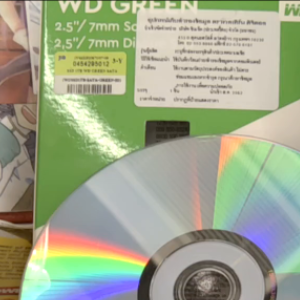}
    \caption*{Ground Truth}
  \end{minipage}

  \caption{Qualitative comparison on the \textit{CD} scene from the Shiny dataset~\cite{wizadwongsa2021nex}. Our method better preserves specular highlights and structural details compared to prior methods.}
  \label{fig:vis_comp_shiny}
\end{figure}

\subsubsection{Datasets}

We evaluate our method on three established benchmarks: NeRF-Synthetic~\cite{mildenhall2020nerf}, Forward-Facing~\cite{mildenhall2019llff}, and Shiny~\cite{wizadwongsa2021nex}.

\subsubsection{Competing Methods}
We compare our method against state-of-the-art neural rendering baselines, grouped into four categories: NeRF variants, voxel/grid-based methods, tensor decomposition approaches, and efficient/explicit representations, plus one additional method. {NeRF Variants:} NeRF \citep{mildenhall2020nerf}, Ref-NeRF \citep{verbin2022ref}, Tetra-NeRF \citep{kulhanek2023tetra}, Point-NeRF \citep{xu2022point}, SRN \citep{sitzmann2019scene}.
 {Voxel/Grid-Based:} Plenoxels \citep{fridovich2022plenoxels}, DVGO \citep{sun2022direct}, PlenOctrees \citep{yu2021plenoctrees}.
 {Tensor Decomposition:} TensoRF (VM-192) \citep{Chen2022ECCV}, K-Planes \citep{fridovich2023k}.
 {Efficient/Explicit Representations:} Instant-NGP \citep{mueller2022instant}, 3D Gaussian Splatting \citep{kerbl3Dgaussians}, Nex \citep{wizadwongsa2021nex}.
 LLFF \citep{mildenhall2019llff}.

Most methods were re-trained under consistent settings for fair comparison, with original results used where applicable.

\section{Experiment Results}

\subsection{Quantitative Results}

Tables~\ref{tab:merged_results} present average quantitative results on NeRF-Synthetic and Forward-Facing datasets under consistent illumination. Our method achieves top performance, surpassing NeRF by 3.17 dB (PSNR) on NeRF-Synthetic and 6.83 dB on Forward-Facing, with superior SSIM and LPIPS scores, indicating enhanced structural and perceptual quality.

Table~\ref{tab:dense_views} shows results for dense-view training on the \textit{materials} scene from NeRF-Synthetic with 100, 200, and 300 views. Our method consistently outperforms baselines, with PSNR improvements of 1.19–3.21 dB over TensoRF, the closest competitor. Notably, our approach scales effectively with view count, unlike 3D GS, which suffers from popping artifacts due to view-dependent effects.

\begin{table}[h]
    \centering
    \caption{Quantitative comparison on NeRF-Synthetic~\citep{mildenhall2020nerf} and Forward-Facing~\citep{mildenhall2019llff}. Best results in \textbf{bold}.}
    \label{tab:merged_results}
    \scriptsize
    \begin{tabular}{lcccccc}
        \toprule
        \multirow{2}{*}{Method} & \multicolumn{3}{c}{NeRF-Synthetic} & \multicolumn{3}{c}{Forward-Facing} \\
        \cmidrule(lr){2-4} \cmidrule(lr){5-7}
        & PSNR$\uparrow$ & SSIM$\uparrow$ & LPIPS$\downarrow$ & PSNR$\uparrow$ & SSIM$\uparrow$ & LPIPS$\downarrow$ \\
        \midrule
        SRN            & 22.26 & 0.846 & 0.170 & --    & --    & --    \\
        NeRF           & 31.01 & 0.947 & 0.081 & 26.50 & 0.811 & 0.250 \\
        LLFF           & --    & --    & --    & 24.41 & 0.863 & 0.211 \\
        PlenOctrees    & 31.71 & 0.958 & 0.053 & --    & --    & --    \\
        Plenoxels      & 31.71 & 0.958 & 0.049 & 26.29 & 0.839 & 0.210 \\
        DVGO           & 31.95 & 0.957 & 0.053 & --    & --    & --    \\
        TensoRF        & 32.39 & 0.957 & 0.057 & 26.51 & 0.832 & 0.217 \\
        Instant-NGP    & 27.22 & --    & 0.206 & --    & --    & --    \\
        $K$-Planes     & 32.37 & 0.962 & 0.150 & 26.80 & 0.846 & 0.193 \\
        3D GS & 32.84 & 0.962 & 0.042 & --    & --    & --    \\
        Tetra-NeRF     & 32.53 & 0.952 & 0.041 & --    & --    & --    \\
        Point-NeRF     & 33.31 & 0.968 & 0.049 & --    & --    & --    \\
        Ref-NeRF       & 33.99 & 0.926 & 0.038 & --    & --    & --    \\
        Nex            & --    & --    & --    & 27.26 & 0.904 & 0.178 \\
        \textbf{Ours}  & \textbf{34.18} & \textbf{0.970} & \textbf{0.037} & \textbf{33.33} & \textbf{0.945} & \textbf{0.117} \\
        \bottomrule
    \end{tabular}

\end{table}

\begin{table}[h]
    \centering
    \caption{Quantitative results on \textit{materials} scene (NeRF-Synthetic) with dense views~\cite{mildenhall2020nerf}.}
    \label{tab:dense_views}
\scriptsize
    \begin{tabular}{clccc}
        \toprule
        \#Views & Method & PSNR$\uparrow$ & SSIM$\uparrow$ & LPIPS$\downarrow$ \\
        \midrule
        \multirow{7}{*}{100} 
        & TensoRF        & {31.42} & {0.961} & {0.020} \\
        & DVGO           & 30.60 & 0.959 & 0.049 \\
        & 3D GS          & 30.49 & 0.960 & 0.037 \\
        & $K$-Planes     & 29.46 & 0.949 & 0.064 \\
        & Instant-NGP    & 25.27 & --    & 0.303 \\
        & Plenoxels      & 29.13 & 0.949 & 0.057 \\
        & \textbf{Ours}  & \textbf{32.61} &\textbf{0.970} & \textbf{0.015} \\
        \midrule
        \multirow{7}{*}{200} 
        & TensoRF        & {32.68} & {0.969} & {0.016} \\
        & DVGO           & 31.63 & 0.967 & 0.045 \\
        & 3D GS          & 27.96 & 0.936 & 0.088 \\
        & $K$-Planes     & 30.75 & 0.961 & 0.053 \\
        & Instant-NGP    & 27.65 & --    & 0.202 \\
        & Plenoxels      & 30.99 & 0.967 & 0.039 \\
        & \textbf{Ours}  & \textbf{34.77} & \textbf{0.979} &\textbf{0.010} \\
        \midrule
        \multirow{7}{*}{300} 
        & TensoRF        & {33.06} & {0.971} & {0.015} \\
        & DVGO           & 31.96 & 0.968 & 0.043 \\
        & 3D GS          & 27.95 & 0.936 & 0.088 \\
        & $K$-Planes     & 30.92 & 0.962 & 0.052 \\
        & Instant-NGP    & 28.12 & --    & 0.167 \\
        & Plenoxels      & 31.22 & 0.969 & 0.038 \\
        & \textbf{Ours}  & \textbf{36.27} & \textbf{0.983} &\textbf{0.009} \\
        \bottomrule
    \end{tabular}
\end{table}

\subsection{Qualitative Results}
Figure~\ref{fig:llffvisually} and Figure~\ref{fig:vis_comp_shiny} showcase qualitative improvements in rendering fine textures (\textit{Horn}, \textit{Fortress}, \textit{Ferns}) and reflective surfaces (\textit{Materials}, \textit{Drums}, \textit{Ficus}, CDs, book covers), validating our method’s ability to model geometry and appearance synergistically.

\subsection{Ablation Study}
\label{sec:ablation}

\begin{table}[t]
  \centering
  \caption{
    {Quantitative results for the ablation study on core components.}  Best results are in \textbf{bold}.
  }
  \label{tab:ablation_components}
    \scriptsize
  \begin{tabular}{llccc}

    \toprule
    \# & Model Variant & PSNR $\uparrow$ & SSIM $\uparrow$ & LPIPS $\downarrow$ \\
    \midrule
    & \textbf{Ours (Full Model)} & \textbf{32.54} & \textbf{0.968} & \textbf{0.041} \\
    (a) & w/o Decomposition & 29.13 & 0.925 & 0.158 \\
    (b) & w/o Dynamic Radiance Integrator & 31.02 & 0.951 & 0.083 \\
    (c) & w/o FiLM in $F_\theta$ & 31.78 & 0.959 & 0.065 \\
    (d) & w/o Neural Basis & 32.21 & 0.964 & 0.052 \\
    \bottomrule
  \end{tabular}
\end{table}

\paragraph{Effect of Dynamic Decomposition.}
In variant (a), we remove our entire dynamic coefficient decomposition module (including $F_\theta$, $H$, and $G_\phi$) and replace it with a standard MLP appearance head that maps appearance features and view direction directly to color. As shown in Table~\ref{tab:ablation_components}, this results in the most significant performance drop among all ablations (over 3dB in PSNR and a nearly 4x worse LPIPS score). Qualitatively, this variant fails to render sharp, view-consistent specular highlights, producing blurry or smeared reflections instead. This strongly confirms that our decomposition framework is fundamental to successfully disentangling material properties from view-dependent effects for high-fidelity rendering.

\paragraph{Analysis of the Blending Mechanism.}
For variant (b), we replace the non-linear Dynamic Radiance Integrator $G_\phi$ with a simple linear operation (a dot product between coefficients and bases). While this model retains the decomposition structure, its performance still drops noticeably, especially on the perceptual LPIPS metric. The rendered results show highlights that, while present, appear dull and lack nuance, failing to capture the rich details arising from complex light-material interactions. This demonstrates that non-linear blending is crucial for synthesizing photorealistic radiance.

\paragraph{Analysis of the Coefficient Network Design.}
In variant (c), we replace the FiLM-like affine transform in the Coefficient Network $F_\theta$ with simple feature concatenation. The results show a measurable decrease in performance. We posit that simple concatenation dilutes the view-direction signal, making it difficult for the network to learn the high-frequency details caused by subtle changes in viewpoint. Our FiLM-inspired design, however, explicitly uses the view direction to modulate other features, enabling more precise control over specular effects.

\paragraph{Effect of the Neural Basis.}
In variant (d), we remove the projection to a neural basis and instead feed the higher-dimensional raw appearance features directly into the Dynamic Radiance Integrator. Interestingly, this variant shows only a slight performance drop. However, this comes at the cost of model efficiency and parameter count. The Dynamic Radiance Integrator must process a much larger input, increasing its complexity. More importantly, distilling features into a compact, shared basis forces the model to learn a reusable dictionary of reflectance patterns, which can lead to better generalization and interpretability. Thus, the neural basis design is key to achieving an optimal balance between performance and efficiency.

\subsection{Analysis of Hyperparameters}

We also analyze the impact of the number of neural bases, $N_p$, on the model's expressive power.

\begin{table}[h]
  \centering
  \caption{
    {Impact of the number of neural bases ($N_p$).} Too few bases limit the model's expressiveness, while too many can lead to diminishing returns and potential overfitting.
  }
  \small
  \label{tab:ablation_hyperparams}
  \begin{tabular}{cc}
    \toprule
    $N_p$ (Number of Bases) & PSNR $\uparrow$ \\
    \midrule
    4 & 31.26 \\
    8 & 32.11 \\
    16 & \textbf{32.54} \\
    32 & 32.57 \\
    \bottomrule
  \end{tabular}
\end{table}

As shown in Table~\ref{tab:ablation_hyperparams}, performance improves significantly as $N_p$ increases from 4 to 16, indicating that more bases allow the model to represent richer material reflectance patterns. However, the improvement becomes marginal when increasing from 16 to 32. This suggests that 16 bases are sufficient for the test scenes, and more may introduce a risk of overfitting and unnecessary computational overhead. We therefore use $N_p=16$ as our default configuration in all main experiments.

\section{Conclusion} \label{Discussion}
In this paper, we introduced a novel neural rendering framework based on {dynamic coefficient decomposition} to address the challenge of rendering highly reflective scenes. Our method models complex appearance by decoupling a static, shared {neural basis} representing material properties from dynamic {coefficients} that are generated by a {Coefficient Network} to adapt to view and illumination. Experiments demonstrate that this approach achieves state-of-the-art quality and quantitative performance, particularly in rendering specular highlights and other view-dependent effects.

While effective, our method's latent-based illumination model is limited in handling local lighting effects like cast shadows, and it does not yet yield editable physical material parameters. Key directions for future work therefore include integrating more structured lighting representations for photorealistic relighting, and extending the framework to predict editable material properties. We believe the decomposition paradigm presented here offers a promising new direction for neural rendering.

\section{Limitations}
\label{sec:limitations}
Our primary limitations are twofold. First, the latent-based illumination model prevents photorealistic relighting with local effects such as cast shadows. Second, our framework does not disentangle explicit physical parameters, thus disallowing direct material editing.

\bibliography{aaai}

\end{document}